\documentclass{article}
\usepackage{amsmath,spconf,amsmath,graphicx,amssymb,multirow,array,booktabs,graphicx,epstopdf,subfigure}
\usepackage[colorlinks]{hyperref}
\newcommand{\keypoint}[1]{\vspace{0.0cm}\noindent\textbf{#1}\quad}

\title{Heterogeneous Domain Generalization via Domain Mixup}
\name{Yufei Wang$^{1,2}$, Haoliang Li$^{2}$, and Alex C. Kot$^{2}$
\thanks{This research was carried out at the Rapid-Rich Object Search (ROSE) Lab, Nanyang Technological University (NTU), Singapore and supported by grants from NTU’s College of Engineering (M4081746.D90). This research was also supported by SSIJRI (Project No.206-A018001). H. Li thanks the Wallenburg-NTU Presidential Postdoc Fellowship grant. }
}
\address{\\
$^{1}$ University of Electronic Science and Technology of China, China\\
$^{2}$ Nanyang Technological University, Singapore}
\begin{document}
%
\maketitle
\begin{abstract}
One of the main drawbacks of deep Convolutional Neural Networks (DCNN) is that they lack generalization capability. In this work, we focus on the problem of heterogeneous domain generalization which aims to improve the generalization capability across different tasks, which is, 
how to learn a DCNN model with multiple domain data such that the trained feature extractor can be generalized to supporting recognition of novel categories in a novel target domain.
To solve this problem, we propose a novel heterogeneous domain generalization method by mixing up samples across multiple source domains with two different sampling strategies. Our experimental results based on the Visual Decathlon benchmark demonstrates the effectiveness of our proposed method. The code is released in  \url{https://github.com/wyf0912/MIXALL}.
\end{abstract}
\section{Introduction}
\label{sec:intro}

Deep Convolutional Neural Networks (DCNN) are widely used in different computer vision tasks (e.g. object recognition/detection, semantic segmentation). However, they are also known to lack generalization capability, especially under the scenario that large-scale training data cannot be obtained. To mitigate such a problem, one can use a DCNN pre-trained by large-scale publicly available data (e.g. ImageNet \cite{Imagenet}), and further finetune the network based on another classification task with few training samples. Another strategy is to directly extract feature representation from a pre-trained DCNN and further use another classifier (e.g. support vector machine, shallow multi-layer network) for recognition purpose. However, directly finetuning the DCNN or extracting features from a pre-trained DCNN may not be able to achieve desired performance mainly due to the domain gap between training and testing data, which further leads to poor generalization capability of trained classifier and brings negative impact on the practical application of DCNN. 

Many efforts have been done to improve the generalization capability of DCNN. Domain adaptation and domain generalization are related research directions, which aim to mitigate the influence of domain shift between training and testing data. Domain adaptation (DA) \cite{DA,DA2,UDA,li2019heterogeneous} assumes that we can obtain the samples from the testing data (a.k.a. target domain) during the training stage, but have no label or few labels. 
Universal domain adaptation (UDA) \cite{UDA} is a relatively new research branch of DA. In UDA, labels are divided into two parts, common label set and private label set. You \textit{et al.} \cite{UDA} adopt the sample-level transferability to develop a weighting mechanism. Another related research direction is partial domain adaptation (PDA) \cite{PDA}, which assumes that the target label space is a subset of the source label space. Compared with the aforementioned two tasks,  open set domain adaptation (OSDA) \cite{osda} is more challenging as it sets private labels as an "unknown" class.  These tasks all assume that there is an intersection between the source and target domain.

Domain generalization (DG) \cite{MMD_AAE,DLOW,crossGrad,Meta-Reg,Reptile,invarientDG} aims to address the harder setting where the target domain is not available during training stage. The general idea of DG is to train the model which is expected to generalize better by only using the data from the source domain(s) and then evaluate based on the ``unseen" target domain. While both DA and DG require that the label space of source and target to be the same,  heterogeneous domain generalization  (HDG) \cite{HDG} assumes that different domains share different label space. In the case of HDG, the label space between source and target are disjoint. Such setting is widely encountered in the field of computer vision where the pre-trained model can be re-used for different types of applications which lack data, computation or human expert time. To address the HDG problem, a feature-critic network with meta-learning mechanism for regularization which was proposed in \cite{HDG}. To simulate the domain shift problem during different stages, the training data were split into two parts, meta-training set and meta-validation set.

In this paper, we aim to tackle the HDG problem by proposing a novel algorithm built upon mixup strategy \cite{mixup} by regularizing the trained model on a convex combination of samples from different domains. In particular, we propose two mixup algorithms. One is to consider the linear behavior in-between training examples, another is to consider the linear relationship among multiple domains. Our experimental results on Visual Decathlon (VD) benchmark shows that our method can outperform state-of-the-art baselines on HDG.

\section{Proposed Method}
\label{sec:format}
\subsection{Problem Definition and Notation}
The goal of HDG is to train a powerful feature extractor which can be used off-the-shelf by a new task. Specifically, for a classification task, we aim to learn a feature extractor $f_\theta$ parameterized by $\theta$ which can be shared across different tasks. Assuming that we have $N$ datasets $\mathcal{D}=\{D_1,D_2,...,D_n\}$, and for each dataset (domain) we have the training data $D_j=\{(x_i^j,y_i^j)\}_{i=1}^{n_j}$, where $n_j$ is the number of data pairs in domain $D_j$, $x_i^j$ is the $i$th data from domain $D_j$ and $y_i^j$ is the corresponding label. To be specific, $y_i^j \in \{0,1,...,k_j\}$ where $k_j$ is the class number in domain $j$. 
We further denote $\hat{y}_i^j$ as the one-hot encoding of $y_i^j$. 
In HDG problem, the label space is disjoint, i.e., for arbitrary $k_1,k_2$, the true class behind $y_{k_1}^i$ is different from that of $y_{k_2}^j$ for $\forall i \neq j$. Regarding the HDG setting, the datasets can be seperated into two parts, $\mathcal{D}_s=\{D_1,D_2,...,D_m\}$, which contains sufficient labeled samples, as source domains for DCNN training purpose, and $\mathcal{D}_t=\{D_{m+1},D_{m+2},..., D_{n}\}$, which contains much fewer labeled samples, as target domains. The object is to train a DCNN $g_{\alpha} \cdot f_{\theta}$, where $g_{\alpha}$ is the classification model parameterized by $\alpha$, on $\mathcal{D}_s$, and further train a shallow model such as SVM or KNN on $\mathcal{D}_t$ with the feature extracted from $f_{\theta}$. We conduct evaluation on testing set of target domain, which draws from the same distribution as $\mathcal{D}_t$.

\subsection{A baseline for HDG}
\label{sec:flatten}
We first consider a simple baseline model called aggregation (AGG) for HDG, which minimize the following loss:
\begin{equation}
    \underset{\theta , \alpha}{min}\sum_{D_j \in \mathcal{D}_s}\sum_{(x_i^j,y_i^j) \in D_j} l(g_\alpha(f_\theta(x_i^j)),\hat{y}_i^j)
\end{equation}
Here $l$ is the cross-entropy loss function. $x_i^j$ is the $i${th} image in domain $D_j$ and $\hat{y}_i^j$ is the one-hot augmented label transferred from the original label $y_i^j$ 
. We can get the length of one-hot vector $\hat{y}_i^j$ by 
\begin{equation}
    |\hat{y}_i^j| = \sum_{D_j \in \mathcal{D}_s}k_j
\end{equation}

Specifically, the relationship between $\hat{y}_i^j$ and $y_i^j$ can be given as
\begin{equation}
    \hat{y}_i^j(n)=
    \left\{\begin{matrix}
    1 &   \quad if \ \ n>k_0 \ and\  y_i^j+\sum_{t=1}^{j-1}k_t = n
    \\
    1 &   \quad if \ \ n<=k_0 \ and\  y_i^j = n
    \\
    0 &  otherwise
    \end{matrix}\right.
\end{equation}
where $\hat{y}_i^j(n)$ is the $n_{th}$ element in one-hot label $\hat{y}_i^j$, $k_t$ is the number of pairs in $\mathcal{D}_t$. 
Then we fix the module $f_\theta$ and use it to extract the features from $\mathcal{D}_t$. We can train a set of classifiers $G=\{g_{\alpha _i}\}_{i=1}^{|\mathcal{D}_t|}$ using the training set of datasets $\mathcal{D}_t$. Finally, we test the classification accuracy of $G$ using the testing set of target domain.

\subsection{MIXUP for HDG}
MIXUP distribution is first proposed by \cite{mixup} which aims to alleviate the behaviors of DCNN such as memorization and sensitivity to adversarial examples. The original mixup distribution is defined as follow

\begin{equation}
\begin{split}
   \mu(\tilde{x}, \tilde{y} | x_i^q, \hat{y}_i^q) = \frac{1}{n} \sum_{j=1}^n \mathbb{E} _{\lambda} [ \delta(\tilde{x} = \lambda x_i^q + (1-\lambda) x_j^q, \tilde{y} = \lambda \hat{y}_i^q +
   (1-\lambda) \hat{y}_j^q) ]
\end{split}
\end{equation}
where $\lambda \in (0,1) $, $\mathbb{E}$ denotes expectation operation,  $\delta$ is Dirac delta function \cite{mixup}.
Here, the data $\{(x_i^q,y_i^q)\}_{i=1}^{n}$ come from the same distribution $q$.

However, in the task of HDG, the training data come from different distributions and disjoint label space. In order to let the mixup distribution fit the HDG task, we proposed a new distribution named HDG mixup distribution which can integrate information from different domains
\begin{equation}
\begin{split}
   \mu_{HDG}(\tilde{x}, \tilde{y} | x_i^q, \hat{y}_i^q)= \frac{1}{n_{sum}} \sum_{j=1}^{|\mathcal{D}_s|}\sum_{k=1}^{|D_j|} \mathbb{E} _{\lambda} [ \delta(\tilde{x}
   = \lambda \cdot x_i^q + (1-\lambda) \cdot x_k^j,
   \tilde{y}= \lambda \cdot \hat{y}_i^q +
   (1-\lambda) \cdot \hat{y}_k^j) ]
\end{split}
\end{equation}
where $n_{sum}$ is the total number of samples in $\mathcal{D}_s$, $(x_i^q,\hat{y}_i^q) $ is a feature-target pair from an arbitrary domain $D_q$ in $\mathcal{D}_s$. 

After encoding the label to a one-hot vector using the method in Sec.\ref{sec:flatten}, we can sample mixed feature-target pairs from the HDG mixup distribution.


In this work, we propose two sampling strategies from MIXUP distribution below.

\keypoint{MIXUP from two domains}
\label{sample-strategy}
The first sampling strategy we proposed is to mix up from two arbitrary domains in one iteration, abbreviated as MIXUP. Specifically, we can sample two data pairs from two arbitrary domains in one iteration and generate a new mixed data pair $(\tilde{x},\tilde{y})$
\begin{equation}
  \tilde{x} = \lambda x_i^p + (1 - \lambda) x_j^q \qquad 
  \tilde{y} = \lambda \hat{y}_i^p + (1 - \lambda)\hat{y}_j^q
\end{equation}
where $\lambda \sim \text{Beta}(\alpha, \alpha)$ for $\alpha \in (0, \infty)$. $p,q$ are discrete random variables draw from $\{1,2,...,|\mathcal{D}_s|\}$. $(x_i^p, \hat{y}_i^p)$ and $(x_j^q, \hat{y}_j^q)$ are two feature-target pairs drawn at
random from two arbitrary domains, and $\lambda \in [0, 1]$ which is controlled by the hyper-parameter $\alpha$.

\keypoint{MIXUP from all domains}
We can also mixup from all domains(MIX-ALL), i.e., mixup all samples from  $|\mathcal{D}_s|$ source domains in one iteration.

\begin{equation}
  \tilde{x} = \sum_{q=1}^{|\mathcal{D}_s|}\lambda_q x^q_i \qquad 
  \tilde{y} = \sum_{q=1}^{|\mathcal{D}_s|}\lambda_q \hat{y}^q_i
\end{equation}

\begin{figure}[ht]
    \centering
    \includegraphics[width=12cm,trim=50 2 50 2,clip]{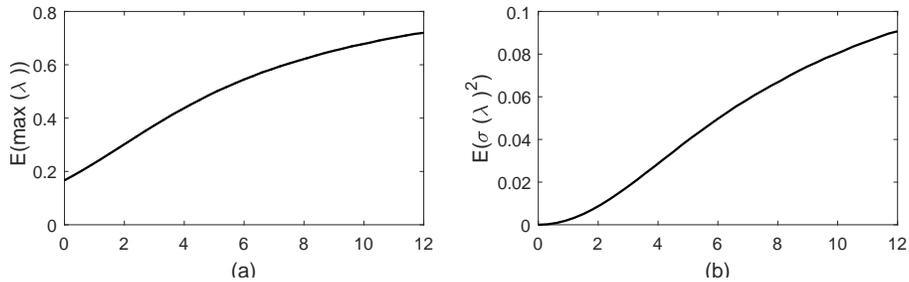}
    \caption{The relationship of hyper-parameter $\beta$ and the expectation of maximum value and variance of random variable vector $\boldsymbol{\lambda}$ under the condition of $|\mathcal{D}_s|=6$}
    \label{fig:VAR-MAX}
\end{figure}

Here $(x_i^q,\hat{y}_i^q)$ is an arbitrary data pair from domain $D_q$, $\tilde{x}$ is the generated sample and $\tilde{y}$ is the corresponding label. $\lambda_q$ is a random variable to control the degree of mixing which is generated and normalized as below
\begin{equation}
\label{uniform}
    \lambda_q =  \frac{e^{u_q}}{\sum_{q=1}^{|\mathcal{D}_s|}e^{u_q}}
\end{equation}
where random variable $u_q\sim U(0,\beta)$. We introduce a hyper-parameter $\beta$ to control the generation of random variable vector $\boldsymbol{\lambda}=[\lambda_1,\lambda_2,...,\lambda_{|\mathcal{D}_s|}]$. The relationship between $\beta$ and the expectation of $max(\boldsymbol{\lambda})$ and $\sigma(\boldsymbol{\lambda})^2$ can be found in Fig.\ref{fig:VAR-MAX}, which shows that with the increase of $\beta$, generated samples become more definiteness and less randomness.

\section{Experiment}
We evaluate our model on Visual Domain Decathlon (VD) benchmark~\cite{VD-score} by following \cite{HDG}.

\begin{table*}[ht]
\centering

\caption{classification accuracy ($\%$) and VD scores on target datasets in VD using SVM.}\label{tab:mainVD}
\label{tab:SVM}
\scalebox{0.9}{
\begin{tabular}{c|ccccccccc}
\toprule
\multirow{2}{*}{Target} & \multicolumn{8}{c}{SVM Classifier}\\
 & Im.N. PT & CrossGrad \cite{crossGrad} & MR \cite{Meta-Reg} & MR-FL \cite{HDG} & Reptile \cite{Reptile} & AGG \cite{HDG} &  FC \cite{HDG}& MIXUP& MIX-ALL\\
\hline
Aircraft & 16.62 & 19.92 & 20.91 & 18.18 & 19.62 & 19.56 & 20.94& \textbf{21.00} &20.61 \\
D. Textures & 41.70 & 36.54 & 32.34 & 35.69 & 37.39 & 36.49 & 38.88& \textbf{39.36}&39.04\\
VGG-Flowers & 51.57 & 57.84 & 35.49 & 53.04 & 58.26 & 58.04 & 58.53& 59.71&\textbf{61.37}\\
UCF101 & 44.93 & 45.80 & 47.34 & 48.10 & 49.85 & 46.98 & 50.82& 51.74&\textbf{55.99}\\
\hline
Ave. & 38.71 & 40.03 & 34.02 & 38.75 & 41.28 & 40.27 & 42.29& 42.95& \textbf{44.25}\\
\hline
VD-Score & 308 & 280 & 269 & 296 & 324 & 290 & 344& 357& \textbf{400}\\
\bottomrule
\end{tabular}
}
\end{table*}

\begin{table*}[ht]
\centering
\caption{classification accuracy ($\%$) and VD scores on target datasets in VD using KNN.}\label{tab:mainVD}
\label{tab:KNN}
\scalebox{0.9}{
\begin{tabular}{c|ccccccccc}
\toprule
\multirow{2}{*}{Target} & \multicolumn{8}{c}{KNN Classifier}\\
 & Im.N. PT & CrossGrad \cite{crossGrad} & MR \cite{Meta-Reg} & MR-FL \cite{HDG} & Reptile \cite{Reptile} & AGG \cite{HDG} &  FC \cite{HDG}& MIXUP&MIX-ALL\\
\hline
Aircraft & 11.46 & 15.93 & 12.03 & 11.46 & 13.27 & 14.03 & \textbf{16.01}& 14.52&14.85 \\
D. Textures & 39.52 & 31.98 & 27.93 & 39.41 & 32.80 & 32.02 & \textbf{34.92}& 34.26&33.67\\
VGG-Flowers & 41.08 & 48.00 & 23.63 & 39.51 & 45.80 & 45.98 & 47.04& 48.24&\textbf{49.71}\\
UCF101 &35.25 & 37.95 & 34.43 & 35.25 & 39.06 & 38.04 & 41.87& 45.03&\textbf{46.41}\\
\hline
Ave. & 31.83 & 33.47 & 24.51 & 31.41 & 32.73 & 32.52 & 34.96& 35.51& \textbf{36.16}\\
\hline
VD-Score & 215 & 188 & 144 & 215 & 201 & 189 & 236& 259&\textbf{268}\\
\bottomrule
\end{tabular}
}
\end{table*}

\keypoint{Dataset.} There are ten heterogeneous domains in VD, including ImageNet \cite{Imagenet}, CIFAR-100 \cite{CIFAR}, Aircraft \cite{aircraft}, Describable textures \cite{texture}, German traffic signs \cite{GTSRB}, etc. 
Different domains contain different image categories as well as a different number of images, e.g., SVHN (Street View House Number) \cite{SVHN} Dataset has 10 classes and 47217 training samples, UCF101 Dynamic Images, a compact representation of videos proposed by \cite{UCF101}, has 101 classes and 7629 training samples, etc. We regard the label space of them disjoint, which is suitable for HDG task. All images in VD have been resized to have a shorter size of 72 pixels.

By following \cite{HDG}, we split VD into two parts: source domain $\mathcal{D}_s$ and target domain $\mathcal{D}_t$. In particular, ${\mathcal{D}_s}$ includes six larger domains, i.e., CIFAR-100 \cite{CIFAR}, Daimler Ped \cite{Daimler}, GTSRB \cite{GTSRB}, Omniglot \cite{Omniglot}, SVHN \cite{SVHN} and ImageNet \cite{Imagenet}. ${\mathcal{D}_t}$ includes four smaller domains, i.e., Aircraft \cite{aircraft}, D. Textures \cite{texture}, VGG-Flowers \cite{vggflower} and UCF101 \cite{UCF101}.

\keypoint{Networks and Training Details.}
We use ResNet-18 \cite{resnet} as the backbone network for all experiments. For fair comparison, we use the same network architecture as \cite{HDG} and also freeze the first four blocks of ResNet-18. And we use the weight pre-trained from aggregation network (AGG) \cite{HDG} for parameter initialization. Noted that there are some differences between the architecture \cite{HDG} with our work. In AGG, each domain has a independent classifier which is an one-layer fully connected layer $g_{\alpha_i}$ where $\alpha_i$ is the parameters of the classifier of domain $i$. The operation of a fully connected layer $g_{\alpha_i}(x)$ can be written in a matrix form given as $g_{\alpha_i}(\mathbf{x}) = \sigma(\mathbf{M}_i\mathbf{x}+\mathbf{b}_i)$ where $\mathbf{M}_i$ is the mapping matrix of fully connected layer with the shape of $(output\_dim,input\_dim)$, $\mathbf{b}_i$ is the bias, which is a column vector of size $output\_dim$, $\mathbf{x}$ is a column input vector, and $\sigma$ is the activation function (e.g. ReLU). To this end, we can adapt AGG in \cite{HDG} into our formulation as
\begin{equation}
    \mathbf{M} = cat(\{\mathbf{M}_i\}_{i=0}^{n}) \qquad \mathbf{b} = cat(\{\mathbf{b}_i\}_{i=0}^{n}) 
\end{equation}
Here $n$ is the number of training domains, $cat$ is the column-level concatenation function. Thus, the classifier weight can be formulated by $\mathbf{M}$ and $\mathbf{b}$. For evaluation, we use the  feature extracted by feature extractor to train an SVM and KNN based on the training set of target domain data, as suggested in \cite{HDG}. We report the classification accuracy and VD-Score \cite{VD-score} for comparison.

Regarding parameter tuning for SVM and KNN, we use a grid-search strategy. We split each target domain into the training part and valid part, then use the cross validation in training set. Finally, we select the parameter group which has the best performance in the training process and use it to validate the quality of the extracted feature.

For implementation, we train the network by using AMSGrad \cite{adam} in an end-to-end manner. We choose the size of minibatch as 64 for each domain. We use an initial learning rate of 0.0005 and a weight decay of 0.0001 for 20k iterations. The learning rate decrease by a factor of 10, 10 after 8k, 16k iterations. We use the mixup strategy during the training phase by choosing $\alpha=0.4$ (which is the hyperparameter of Eq.(\ref{eq.mixup}) to control the strength of mixup) in the first strategy and $\beta=6$ in the second sampling strategy.

\keypoint{Results.}
We evaluate our proposed method by using the VD benchmark and compare against different state-of-the-art DG baselines, including MetaReg(MR) \cite{Meta-Reg}, CrossGrad \cite{crossGrad}, and Reptile \cite{Reptile}. We also compare with Feature-Critic Networks(FC) \cite{HDG} as another baseline, which was the first work considering HDG problem. 

We first evaluate by considering SVM for classification purpose. The results are shown in Table \ref{tab:SVM}. Im.N.PT in Table \ref{tab:SVM} denotes the result by extracting features with the ImageNet pre-trained resnet18 model. Based on the results, we find that our proposed method can significantly outperform all other baselines based on classification accuracy as well as VD Score \cite{VD-score}, which reflects whether the method can achieve stable performance or not. 

Then we evaluate our proposed method by considering KNN as classifier.  The results are shown in Table \ref{tab:KNN}. We find that we can also achieve much better performance compared with all other baseline methods. 

\begin{figure}[ht]
    \centering
    \includegraphics[width=11cm, trim=50 0 50 2,clip]{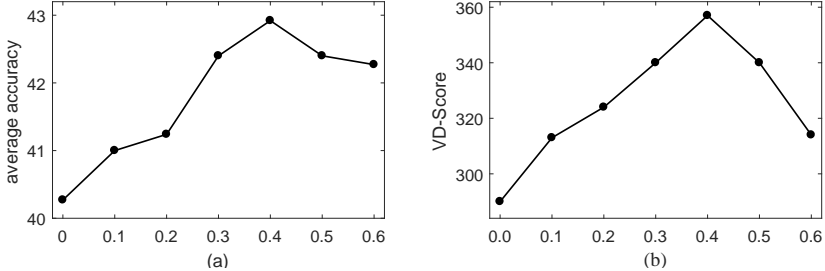}
    \caption{The performance of MIXUP with different $\alpha$ which controls the strength of mixup. }
    \label{fig:mixup1}
\end{figure}

\begin{figure}[ht]
    \centering
    \includegraphics[width=11cm, trim=50 3 160 76,clip]{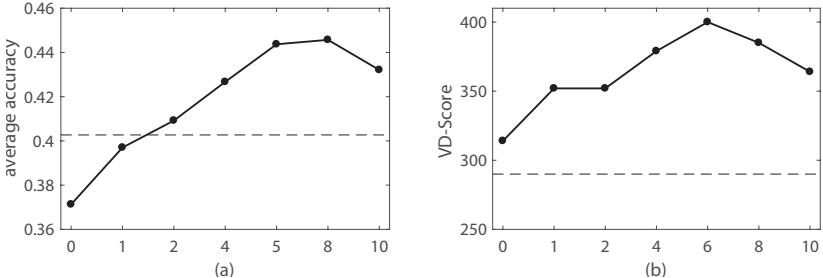}
    \caption{The performance of MIX-ALL with different hyperparameter $\beta$ which controls the strength of mixup. In this figure, the dotted line represents the baseline in Sec. \ref{sec:flatten}}
    \label{fig:mixall}
\end{figure}

\keypoint{Ablation Study.} 
To evaluate the influence of parameter setting,  we further conduct experiments with different hyper-parameter in two sampling strategies proposed in  \ref{sample-strategy}.  The results are shown in Fig. \ref{fig:mixup1} and Fig.\ref{fig:mixall} respectively.

Based on Fig.\ref{fig:mixup1}, we observe several interesting findings. First, when the hyperparameter $\alpha$ is close to zero, the model degenerates into the baseline model we proposed as $\lambda$ centers on either 0 or 1.
Second, we find a trend in both accuracy and VD-score that the performance of model improves first, and then drops. We conjecture the reason why performance rises may profit from the enhanced robustness and drop due to underfitting.

We also conduct experiments to study the impact of hyper-parameter $\beta$ which influence the strength of mixup in Eq.(\ref{uniform}). The results are shown in Fig.~\ref{fig:mixall}. We can find that average accuracy has a drop when $\beta$ is equal to 0 or 1. Such result is consistent to our argument that when we choose a small $\beta$ value, the expectation of $max(\boldsymbol{\lambda})$ and $\sigma(\boldsymbol{\lambda})^2$ tend to be small, which may lead to too much noise. When $\beta$ is very large, the model will degenerate into the baseline model as shown in Fig.\ref{fig:VAR-MAX}. We can find that the performance continues to drop when $n$ get larger.

\section{conclusion}
We proposed a novel heterogeneous domain generalization method by considering data augmentation through domain mixup. In particular, we proposed two strategies, mixup between two domains and across multiple domains.
By sampling the augmented training data from the heterogeneous mixup distribution, we got a more robustness feature extractor, which can be greatly helpful when used off-the-shelf. 
The experimental results demonstrate that our proposed methods can outperform all existing state-of-the-art baselines in terms of classification accuracy and VD-score.

\clearpage
\vfill\pagebreak

\bibliographystyle{IEEEbib}
{\small
\bibliography{strings,refs}
}
\end{document}